%% file: 0_main.tex
\documentclass[letterpaper]{article} % DO NOT CHANGE THIS
\usepackage{aaai23}  % DO NOT CHANGE THIS
\usepackage{times}  % DO NOT CHANGE THIS
\usepackage{helvet}  % DO NOT CHANGE THIS
\usepackage{courier}  % DO NOT CHANGE THIS
\usepackage[hyphens]{url}  % DO NOT CHANGE THIS
\usepackage{graphicx} % DO NOT CHANGE THIS
\usepackage{natbib}  % DO NOT CHANGE THIS AND DO NOT ADD ANY OPTIONS TO IT
\usepackage{caption} % DO NOT CHANGE THIS AND DO NOT ADD ANY OPTIONS TO IT
\usepackage{algorithm}
\usepackage{algorithmic}

\usepackage{newfloat}
\usepackage{listings}
\DeclareCaptionStyle{ruled}{labelfont=normalfont,labelsep=colon,strut=off} % DO NOT CHANGE THIS
\floatstyle{ruled}
\newfloat{listing}{tb}{lst}{}
\floatname{listing}{Listing}
\usepackage{marvosym} %add by zhuhao, for envelope symbo

\usepackage{booktabs,caption} % used for table, zhuhao
\usepackage{xcolor,colortbl}% for color a cell in Table, added by zhuhao

\newcommand{\ca}{\cellcolor[rgb]{0.4,1.0,0.8}}
\newcommand{\cb}{\cellcolor[rgb]{0.55,1.0,0.8}}
\newcommand{\cc}{\cellcolor[rgb]{0.7,1.0,0.8}}
\newcommand{\cd}{\cellcolor[rgb]{0.85,1.0,0.8}}
\newcommand{\ce}{\cellcolor[rgb]{1.0,1.0,0.8}}

\title{RAFaRe: Learning Robust and Accurate Non-parametric 3D Face Reconstruction\\ from Pseudo 2D\&3D Pairs}

\author {
    Longwei Guo,
    Hao Zhu$\textsuperscript{\Letter}$,
    Yuanxun Lu,
    Menghua Wu,
    Xun Cao
}
\affiliations {
    Nanjing University, Nanjing, China\\
    \{guolongwei, luyuanxun, menghuawu\}@smail.nju.edu.cn, \{zhuhaoese, caoxun\}@nju.edu.cn
}

\usepackage{bibentry}
\begin{document}

%====================with teaser image====================
\twocolumn[{%
\renewcommand\twocolumn[1][]{#1}%

\maketitle
\vspace{-0.4in}
\begin{center}
    \includegraphics[width=1.0\linewidth]{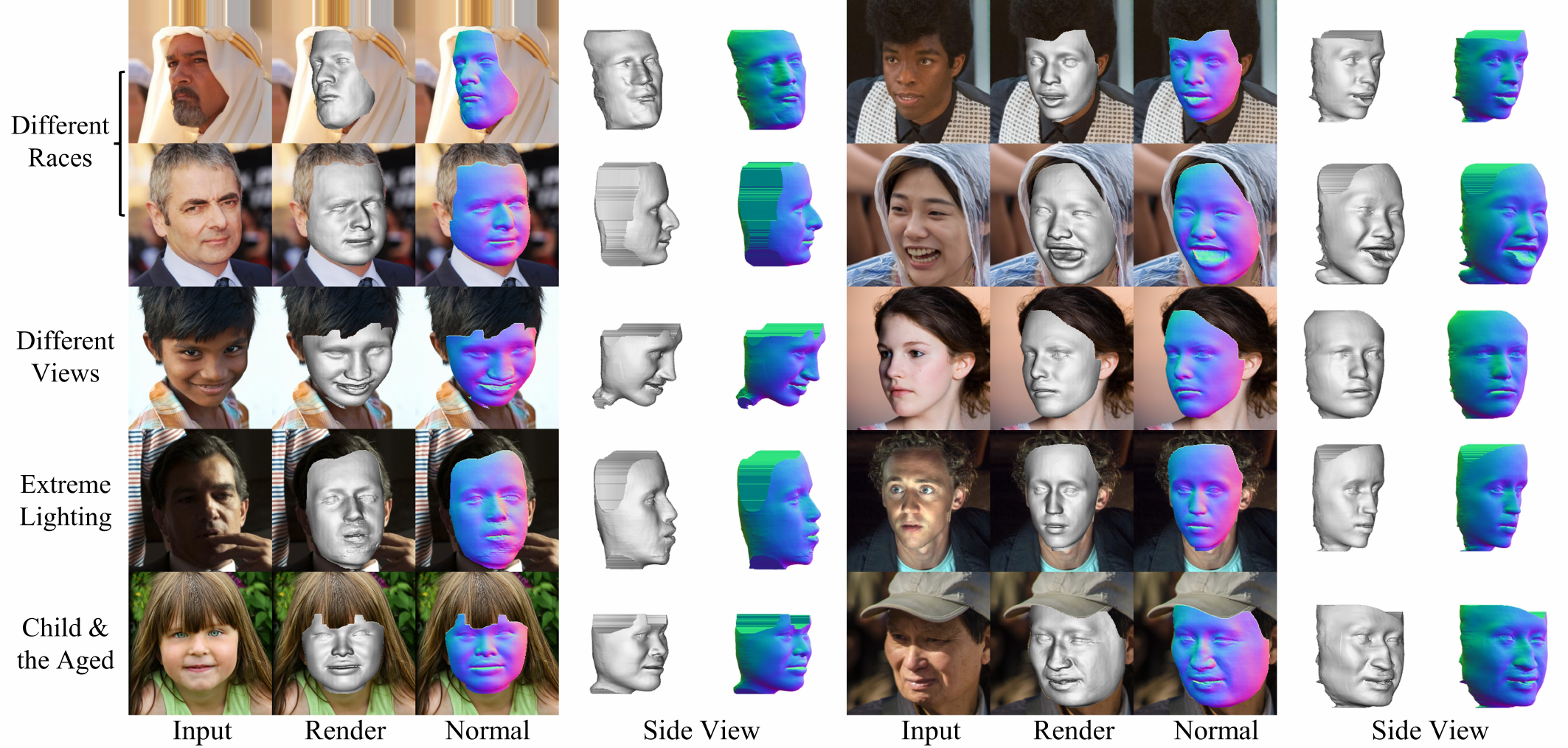}
    \vspace{-0.2in}
    \captionof{figure}{Our method reconstructs high-fidelity and accurate geometry that is generalized for different races, views, lighting and ages. We recommend watching the \emph{supplementary video} for more results.}
\label{fig:show}
\end{center}
}]

%\maketitle

\begin{abstract}
We propose a robust and accurate non-parametric method for single-view 3D face reconstruction (SVFR). While tremendous efforts have been devoted to parametric SVFR, a visible gap still lies between the result 3D shape and the ground truth. We believe there are two major obstacles: 1) the representation of the parametric model is limited to a certain face database; 2) 2D images and 3D shapes in the fitted datasets are distinctly misaligned. To resolve these issues, a large-scale pseudo 2D\&3D dataset is created by first rendering the detailed 3D faces, then swapping the face in the wild images with the rendered face. These pseudo 2D\&3D pairs are created from publicly available datasets which eliminate the gaps between 2D and 3D data while covering diverse appearances, poses, scenes, and illumination. We further propose a non-parametric scheme to learn a well-generalized SVFR model from the created dataset, and the proposed hierarchical signed distance function turns out to be effective in predicting middle-scale and small-scale 3D facial geometry. Our model outperforms previous methods on FaceScape-wild/lab and MICC benchmarks and is well generalized to various appearances, poses, expressions, and in-the-wild environments. The code is released at \url{https://github.com/zhuhao-nju/rafare}.
\end{abstract}

\let\thefootnote\relax\footnotetext{\hspace{-0.22in}$\textsuperscript{\Letter}$ Hao Zhu is the corresponding author. }

%******************BODY TEXT
\input{1_intro.tex}

\input{2_related.tex}

\input{3_method.tex}

\input{4_exp.tex}
\input{5_con.tex}

\clearpage

\section{Acknowledgements}
This work was supported by the NSFC grant 62001213, 62025108, the National Key R\&D Program of China grant 2022YFF0902401, and gift funding from Huawei Research and Tencent Rhino-Bird Research Program.

\bibliography{aaai23}

\end{document}

%% file: 1_intro.tex
\section{Introduction}
\label{sec:intro}

Reconstructing a 3D face shape from an arbitrary single image is a long-standing problem in the computer vision and computer graphics community. 
It attracts much more attention in recent years for its fundamental impact on comprehensive downstream applications, e.g., facial editing, visual effects, facial animation, virtual make-up, and VR/AR character creation.
About two decades ago, Blanz et al. \cite{blanz1999morphable} pioneered the 3D morphable model (3DMM) to tackle this problem, which further developed into a classic paradigm to alleviate the ambiguity of the problem. Many researchers employ a prior parametric model \cite{cao2013facewarehouse,li2017learning,yang2020facescape} as a coarse 3D model and optimize the parameters by minimizing semantic energy functions \cite{cao20133d,thies2016face2face}. In recent years, the rapid development of deep learning pushes the field forward. Deep neural networks are proven to be more effective in regressing the 3DMM parameters, which can be trained on large-scale fitted or synthetic datasets.

Despite the recent advances, achieving \emph{robust} and \emph{accurate} single-view 3D face reconstruction remains an open challenge. The difficulty is two-fold, lying in both face representations and training data. On the one hand, the representation ability of 3DMMs or other parametric models is limited to the fixed sub-linear space of the face database, which makes it hard to recover the fine-scale facial details and out-of-domain face attribute distributions. 
On the other hand, inaccurate and limited training databases restrict the reconstruction performance of both generalization and precision. Concretely, though accurate but limited studio-captured data is available~\cite{yang2020facescape,Dai2019}, it is almost infeasible to obtain large-scale in-the-wild data at the same accuracy, which constrains the model generalization ability to a great extent. Previous approaches~\cite{feng2018joint} tend to adopt in-the-wild image datasets with fitted 3DMM models as training data~\cite{zhu2016face}, failing to reconstruct middle-scale 3D geometry.

In this paper, we create pseudo 2D\&3D datasets from publicly available datasets to address these challenges. Specifically, we propose 1) a novel pipeline for large-scale in-the-wild 2D\&3D facial data generation, and 2) a non-parametric method tailored to recover the facial geometry from a single image. Our reconstruction approach obviates the dependence on prior parametric models and allows for good generalization on poses, expressions, scenes, and illumination.
The pipeline is designed to exploit the data augmentation insights and make the most effective use of existing large-scale in-the-wild facial image datasets and limited lab-environmental precise 3D facial shape datasets. 
Going a step further, we develop a non-parametric method to reconstruct facial geometry from monocular images directly, breaking the limitations of parametric models. 
The most notable improvement is the hierarchical signed distance function to recover detailed facial geometry at different levels. The non-parametric model is trained on the proposed large-scale faithful dataset and shows effective and superior performance over previous state-of-the-art approaches. In summary, we highlight the following contributions:

\begin{itemize}
    \item We introduce a novel pipeline to generate a \emph{large-scale}, \emph{strictly-aligned}, and \emph{in-the-wild} facial dataset with corresponding faithful geometry ground truth. This pipeline enables a higher ceiling for current non-parametric single-view face reconstruction models.
    \item We propose a hierarchical implicit function-based architecture to estimate the detailed 3D face geometry from a single image. The key point of the architecture is the non-parametric model, and unlike most previous 3DMM-based approaches, it is robust to variations of human races, poses, scenes, and illumination.
    \item We achieve the state-of-the-art accuracy on FaceScape-wild/lab \cite{yang2020facescape} and MICC Florence \cite{micc} benchmarks for single-view face reconstructions. Comprehensive qualitative and quantitative evaluations demonstrate the superiority of our method over competing approaches.
\end{itemize}

%% file: 2_related.tex
\section{Related Work}
\label{sec:related}

Single-View 3D Face Reconstruction (SVFR) has been a hotspot for decades due to the enormous significance of potential applications and the intrinsic ambiguity and difficulty compared with other input modalities, e.g., video sequences, RGB-D data, and multi-view images. In this section, we briefly review the prior work in this field, along with related data augmentation methods. We recommend checking the survey \cite{zollhofer2018state} for a comprehensive overview.

\noindent \textbf{Parametric SVFR Methods.} Parametric methods estimate the facial shape by regressing the parameters of 3DMM, which is a statistical model to transform the shape and texture of the faces into a vector space representation \cite{blanz1999morphable}. For the formulation of 3DMM, please check the recent survey \cite{egger20203d}. Traditionally, these methods follow an analysis-by-synthesis schedule and build semantic correspondences between images and statistical models by optimization-based \cite{romdhani2005estimating,amberg2007optimal,zhu2016face,thies2016face2face,dou2017end} or learning-based algorithms \cite{sanyal2019learning,tran2019towards,gecer2019ganfit,tu20203d,koizumi2020look,shang2020self,guo2020towards,deng2019accurate,liu2018disentangling,shang2020self}. Recently, researchers discover a self-supervised scheme to train the models by employing differentiable renderers \cite{tewari2018self,tewari2017mofa,genova2018unsupervised,tan2020self,feng2021learning,deng2019accurate,sanyal2019learning}. These methods inherit the limitations of the 3DMM, which lie in a fixed sub-linear space and generate only coarse shapes without facial details. Several works propose non-linear 3DMMs \cite{tran2018nonlinear,tran2019towards,yenamandra2021i3dmm,zhuang2022mofanerf} to break the traditional limitations.

To add detailed facial shape to the coarse 3DMMs, some methods \cite{yang2020facescape,feng2021learning,chen2019photo,chen2020self} predict displacement maps over the coarse models to represent details. Another attempts \cite{jiang20183d,sengupta2018sfsnet,richardson2017learning,tran2018extreme,zhu2019detailed,zhu2021detailed,riviere2020single} leverage the shape-from-shading method to reconstruct shape details. However, it is error-prone for outer occlusions, specular highlights, and strong cast shadows.

\begin{figure*}[t]
\begin{center}
\includegraphics[width=0.95\linewidth]{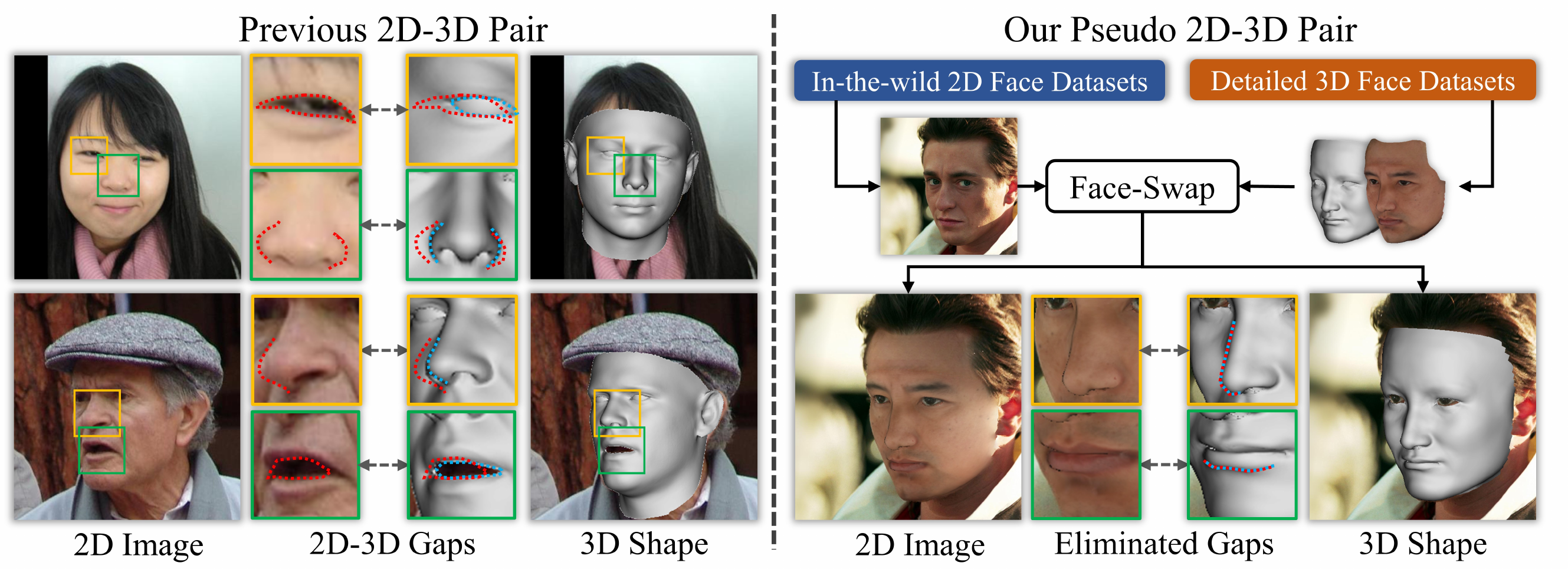}
\vspace{-0.1in}
\end{center}
\caption{Left: In 300W-LP\cite{sagonas2016300} dataset, the middle-scale features in the 2D images and 3D models are mismatched. The zoom-in figure shows that the outlines of the image (red dotted line) and that of the 3D shape (blue dotted line) are obviously misaligned. These 2D\&3D gaps \emph{commonly} exist in previous 2D\&3D face datasets which limited the ceiling for previous reconstruction models. Right: our pseudo-2D\&3D pairs (right) eliminate such gaps.}
\label{fig:gap}
\end{figure*}

\noindent \textbf{Non-parametric SVFR Methods.} Non-parametric methods discard prior models and estimate facial geometry directly. These methods recover 3D faces in the form of volumes \cite{jackson2017large}, meshes \cite{feng2018joint,ruan2021sadrnet,zeng2019df2net,sela2017unrestricted,zhou2019dense,zhu2020beyond}, or depth maps \cite{zhang2021learning}, and hence can capture finer shape details compared with parametric methods. However, the free-form representation leads to fewer constraints and makes it difficult for neural networks to predict the accurate shape. Unlike these approaches, we propose a hierarchical signed distance function-based model to reconstruct coarse- and fine-level geometry in different stages. This representation shows higher robustness, accuracy, and generalization over other non-parametric formulations and overcomes the flaws of 3DMMs.

\noindent \textbf{Synthetic Data Augmentation.} Datasets play a crucial role in the task of learning-based SVFR because their quality decides the performance of neural networks, while current public datasets cannot meet the demands. High-quality scans or multi-view reconstruction systems provide detailed and accurate shapes but lack generalization due to the complexity of the capture system \cite{xiao2022detailed}. On the other hand, in-the-wild datasets \cite{zhu2016face} with optimization-based 3DMM fitting satisfy generality but introduce misalignment between 3D models and 2D images. Methods trained with these datasets \cite{feng2018joint,jackson2017large,guo2020towards} therefore can only recover coarse geometry. Some methods \cite{dou2017end,richardson20163d,varol2017learning,zhu2018view} generate synthetic data for training, but their image is not photo-realistic and the 3D model is not precise enough. Instead, we utilize both accurate but limited lab-captured data and countless unconstrained images and then take the best of both sides. Our face-swapping-based pipeline is able to generate innumerable faithful pseudo-2D\&3D pairs, meaning the network can be trained with diverse and reliable data to attain better performance over previous schemes.

%% file: 3_method.tex
\section{Method}
\label{sec:method}

\subsection{Revisit 2D\&3D Gap}

The quality of 2D\&3D data for training has a crucial impact on the performance of the SVFR model but remains to be a weak point. To achieve optimal performance, the 2D\&3D data for training should be 1) in large amounts; 2) covering diverse facial shapes, appearances, expressions, and environments; and 3) well aligned and accurately corresponded between 2D\&3D. However, these requirements can hardly be met in practice, as capturing accurate 3D facial shapes in the wild is extremely difficult and expensive. A compromised solution that is commonly used is to build 2D\&3D pairs by fitting a 3D face model to the in-the-wild images\cite{sagonas2016300}, as shown in Figure~\ref{fig:gap}. Though the requirements of amount and image diversity are met, the fitted 3D shapes and 2D images are visibly misaligned, leading to a poor fidelity of the predicted facial shape.

\subsection{Pseudo 2D\&3D Pairs}
\label{sec:pseudo}

\begin{figure*}[t]
\begin{center}
\includegraphics[width=1.0\linewidth]{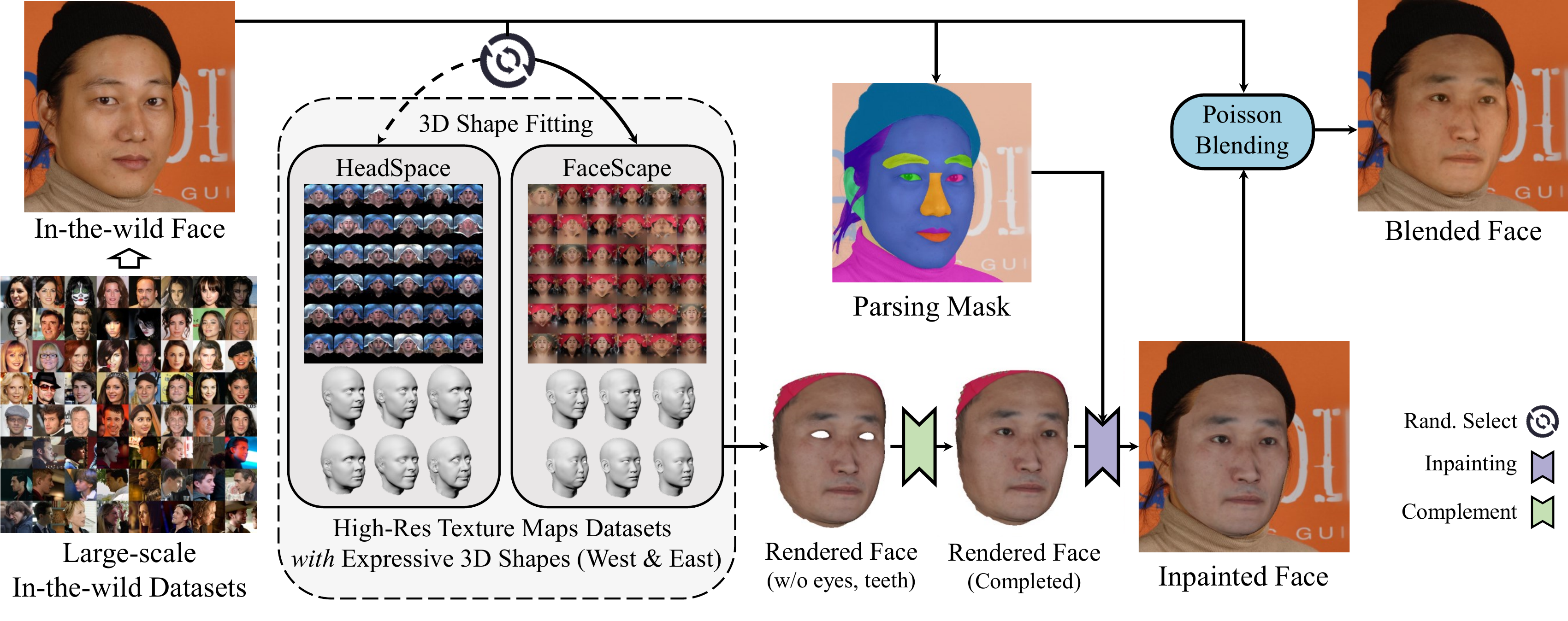}
\vspace{-0.35in}
\end{center}
\caption{Overview of the proposed pseudo 2D\&3D pair synthesis pipeline. }
\vspace{-0.2in}
\label{fig:data_pipeline}
\end{figure*}

\begin{figure}[th]
\begin{center}
\includegraphics[width=0.95\linewidth]{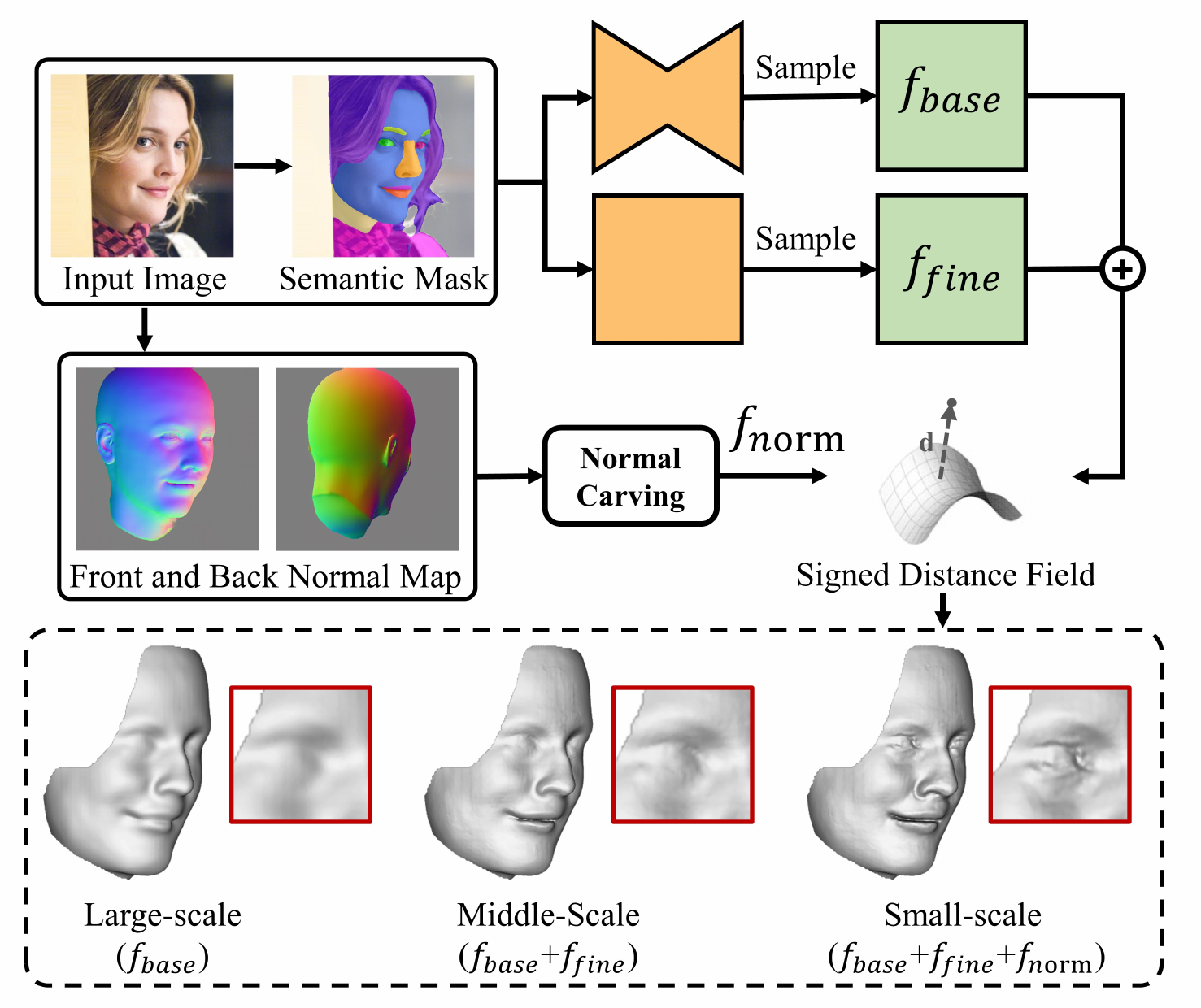}
\vspace{-0.1in}
\end{center}
\caption{Overview of the hierarchical SDF-based network. }
\vspace{-0.2in}
\label{fig:hsdf}
\end{figure}

To eliminate the 2D\&3D gaps, we propose to create pseudo 2D\&3D pairs by swapping in-the-wild faces with accurately reconstructed faces.
As shown in Figure~\ref{fig:data_pipeline}, the pipeline consists of three stages.

In the first stage, we firstly select a face image from the large-scale in-the-wild datasets (CelebA\cite{lee2020maskgan}, FFHQ\cite{ffhq}, IMDB-WIKI\cite{rothe2015dex}), then randomly fit a 3DMM created by FaceScape dataset\cite{yang2020facescape} or HeadSpace dataset\cite{Dai2019} to this image by optimizing 2D facial landmark locations\cite{yang2020facescape}. 
Then, a UV texture map is randomly selected from the two 3D face datasets, and a 2D face can be rendered from the fitted 3D mesh and the selected UV texture map. 

In the second stage, the parsing mask is extracted from the in-the-wild image using BiSeNet\cite{yu2018bisenet}. Then the complementing network synthesizes the missing eyes and teeth for the rendered face, and the inpainting network adjusts the edge of the rendered image to match the parsing mask. Specifically, the face region of the parsing mask and rendered face are concatenated as input of the inpainting network, which aims to obtain the complete face of the intersection region of the facial mask and rendered face. Meanwhile, we use the image warping algorithm to adjust the edge of the in-the-wild image to fit the intersection region. The adjustment aims to make the in-the-wild image and the inpainted image completely aligned. 

In the final stage, Poisson blending is used to merge the in-the-wild face and the inpainted face, generating the blended face. The blended face and the fitted 3D model constitute the pseudo 2D\&3D pairs, which are accurately corresponded and cover diverse facial shapes, expressions, poses, scenes and illuminations. 

In practice, we synthesize $100k$ pseudo 2D\&3D pairs, of which $18k$ images contain large-pose faces. Both our complementing network and inpainting network adopt pix2pixHD~\cite{wang2018pix2pixHD} as the backbone. Please refer to the supplementary material for implementation details.

\begin{table*}[t]
\centering
\caption{Quantitative evaluation on FaceScape in-the-wild benchmark}
\vspace{-0.1in}
\resizebox{\textwidth}{!}
{ 
\begin{tabular}{lccccccccccccc}
\toprule
                            \multicolumn{1}{c}{Pose Angle $\rightarrow$}
                                       & 
                            \multicolumn{3}{c}{$0^\circ-5^\circ$}             & \multicolumn{3}{c}{$5^\circ-30^\circ$}            & \multicolumn{3}{c}{$30^\circ-60^\circ$}            & \multicolumn{3}{c}{$60^\circ-90^\circ$}             &           \\
\multicolumn{1}{c|}{Method \ $\downarrow$} & \,CD\,   & \,MNE\,   & \multicolumn{1}{c|}{\,CR\,} & \,CD\,   & \,MNE\,   & \multicolumn{1}{c|}{\,CR\,} & \,CD\,   & \,MNE\,   & \multicolumn{1}{c|}{\,CR\,} & \,CD\,    & \,MNE\,   & \multicolumn{1}{c|}{\,CR\,} & Succ. \\ \midrule
Ext3dFace~\cite{tran2018extreme}   & 5.03    & 0.158     & 61.5    & 5.52    & 0.176     & 55.7    & 7.92    & 0.208     & 40.4    & 25.39   & 0.266     & 27.1    & 85.5      \\
PRNet~\cite{feng2018joint}           & 2.61    & 0.119     & 83.0    & 3.11    & 0.114     & 82.7    & 4.26    & 0.119     & 78.2    & \cb3.88    & \ce0.140     & \cd75.3    & \ca100.0     \\
D3DFR~\cite{deng2019accurate}   & \cc2.30    & \cb0.070     & 83.1    & \cb2.50    & \cb0.072     & 83.0    & 3.57    & \cd0.082     & 77.8    & 6.81    & 0.143     & 62.4    & \ca100.0     \\
RingNet~\cite{sanyal2019learning}         & \cd2.40    & 0.085     & \cb99.8    & 2.99    & 0.085     & \cb99.7    & 4.78    & 0.100     & \cb98.4    & 10.71   & 0.190     & \cb97.1    & \ca100.0     \\
DFDN~\cite{chen2019photo}            & 3.66    & 0.090     & 86.6    & 3.27    & 0.091     & 86.5    & 7.29    & 0.128     & 84.3    & 27.48   & 0.302     & 57.2    & 88.2      \\
DF2Net~\cite{zeng2019df2net}          & 2.92    & 0.121     & 57.1    & 4.21    & 0.128     & 55.3    & 6.55    & 0.159     & 46.3    & 19.76   & 0.309     & 30.8    & 68.8      \\
UDL~\cite{chen2020self}             & \cb2.27    & 0.091     & 69.0    & \cc2.59    & 0.092     & 68.3    & \ce3.46    & 0.106     & 65.0    & 6.32    & 0.176     & 49.0    & 86.2      \\
FaceScape~\cite{yang2020facescape} & 2.81    & 0.086     & 83.7    & 3.17    & 0.092     & 82.0    & 4.09    & 0.108     & 79.0    & 6.57    & 0.162     & 67.9    & 96.0      \\
MGCNet~\cite{shang2020self}          & 2.97    & \cc0.073     & 84.4    & 2.94    & \cc0.073     & 84.5    & \cb2.78    & \ca0.070     & 81.6    & \cc4.21    & \ca0.091     & 74.3    & \ca100.0     \\
3DDFA\_v2~\cite{guo2020towards}       & \ce2.49    & \cd0.074     & 86.5    & 2.66\ce    & \cd0.074     & 86.0    & \cd3.18    & \cc0.078     & 83.1    & \ca3.67    & \cb0.093     & \ce79.9    & \ca100.0     \\
SADRNet~\cite{ruan2021sadrnet}         & 6.60    & 0.113     & \ce90.2    & 6.87    & 0.113     & \ce89.4    & 6.40    & 0.103     & \ce84.4    & 8.63    & 0.163     & \cd82.7    & \ca100.0     \\
LAP~\cite{zhang2021learning}             & 4.19    & 0.111     & \cd93.5    & 4.47    & 0.116     & \cd92.8    & 6.16    & 0.148     & \cd87.3    & 13.71   & 0.205     & 68.1    & \ca100.0     \\
DECA~\cite{feng2021learning}            & 2.88    & \ce0.080     & \ca99.9    & \cd2.64    & \ce0.079     & \ca99.9    & \cc2.88    & \cd0.082     & \ca99.8    & \cd4.83    & \cc0.116     & \ca99.7    & \ca100.0     \\ 
Ours  & \ca1.79    & \ca0.058     & \cc99.5    & \ca1.92    & \ca0.062     & \cc99.3    & \ca2.12    & \cb0.071     & \cb98.4    & \ce5.26    & \cd0.123    & \cc96.7     & \ca100.0\\
\bottomrule
\end{tabular}
}
\footnotesize{
\leftline{\quad\quad The highest 5 scores are colored: `\colorbox[rgb]{0.4,1.0,0.8}{green}$\rightarrow$\colorbox[rgb]{1.0,1.0,0.8}{yellow}' means `high$\rightarrow$low', similarly hereinafter.}}
\label{tab:eval_wild}
\vspace{-0.1in}
\end{table*}

\begin{table*}[t]
\centering
\caption{Quantitative evaluation on FaceScape in-the-lab benchmark}
\vspace{-0.1in}
\begin{tabular}{lcccccccccc}
\toprule
                            \multicolumn{1}{c}{Pose Angle $\rightarrow$}  &      & $0^\circ$ &                         &      & $30^\circ$ &                         &       & $60^\circ$ &                         &           \\
\multicolumn{1}{c|}{Method \ $\downarrow$} & {CD}   & {\,MNE\,}    & \multicolumn{1}{c|}{\, CR\,} & {\, CD}   & {\, MNE}     & \multicolumn{1}{c|}{\, CR\,} & {\, CD}    & {\, MNE}     & \multicolumn{1}{c|}{\, CR\,} & Succ. \\ \midrule
Ext3dFace~\cite{tran2018extreme}   & 4.59    & 0.131     & 86.2    & 7.42    & 0.170     & 69.1    & 8.51    & 0.175     & 55.2    & 85.9      \\
PRNet~\cite{feng2018joint}           & \cb2.94    & 0.133     & 92.5    & \ca3.40    & 0.125     & 90.1    & \cb3.74    & \ce0.122     & 85.2    & \ca100.0     \\
D3DFR~\cite{deng2019accurate}   & 3.99    & 0.106     & 87.6    & 5.90    & 0.120     & 81.3    & 5.55    & 0.137     & 75.3    & 98.9      \\
RingNet~\cite{sanyal2019learning}         & 3.62    & 0.102     & \ca99.9    & 5.03    & \ce0.111     & \ca99.7    & 6.82    & 0.151     & \cc94.5    & \ca100.0     \\
DFDN~\cite{chen2019photo}            & 4.28    & 0.111     & \cd98.4    & 6.71    & 0.132     & \cd95.2    & 23.63   & 0.280     & 81.0    & 94.7      \\
DF2Net~\cite{zeng2019df2net}          & 4.48    & 0.152     & 64.1    & 7.64    & 0.200     & 52.2    & ---$^*$   & ---$^*$    & ---$^*$  & 73.6      \\
UDL~\cite{chen2020self}             & \ca2.21    & \cb0.092     & 79.5    & 5.34    & 0.123     & 71.3    & 5.63    & 0.167     & 61.9    & 87.0      \\
FaceScape~\cite{yang2020facescape} & \ce3.21    & \ca0.090     & 94.2    & 4.87    & 0.119     & 86.2    & 4.68    & 0.146     & 81.7    & 92.0      \\
MGCNet~\cite{shang2020self}          & 3.45    & 0.085     & 92.7    & \cd3.91    & \ca0.093     & 90.1    & \ca3.65    & \ca0.090     & 83.2    & \ca100.0     \\
3DDFA\_v2~\cite{guo2020towards}       & \cd3.05    & \cc0.093     & 95.2    & \cb3.41    & \cb0.096     & 93.8    & \cc3.82    & \cb0.097     & \cd88.2    & \ca100.0     \\
SADRNet~\cite{ruan2021sadrnet}         & 4.25    & 0.109     & 95.8    & 7.07    & 0.137     & \ce94.9    & 7.09    & 0.148     & \ce87.6    & \ca100.0     \\
LAP\cite{zhang2021learning}             & 4.27    & 0.112     & \ce96.4    & 7.33    & 0.149     & 93.2    & 8.70    & 0.195     & 85.6    & 99.2      \\
DECA~\cite{feng2021learning}            & 3.30    & \cc0.093     & \cb99.8    & \ce4.14    & \cc0.100     & \cb99.4    & \cd4.20    & \cc0.107     & \cb97.1    & \ca100.0     \\
Ours             & \cc2.95    & \cc0.093     & \cc98.9    & \cc3.69    & \cd0.102     & \cc98.7    & \ce5.28    & \cd0.111     & \ca98.3    & \ca100.0      \\
\bottomrule
\end{tabular}
\footnotesize{
\leftline{\quad\quad $^*$ --- means no valid results are generated in this category.}}
\label{tab:eval_lab}
\vspace{-0.1in}
\end{table*}

\subsection{Hierarchical Signed Distance Function}

The creation of pseudo 2D\&3D pairs makes it possible to train a non-parametric SVFR model and break through the constraints of 3DMM solution space. We propose a hierarchical signed distance function to serve as the base of our non-parametric SVFR network, as shown in Figure~\ref{fig:hsdf}.  Inspired by PIFu~\cite{saito2019pifu}, our network represents 3D facial shapes by learning a pixel-aligned signed distance function. The source image and the parsed semantic mask~\cite{yu2018bisenet} are fed into two feature extractors, then two MLPs take pixel-aligned features and depth value $z$ as input, predicting a signed distance for each sample point.  The final predicted signed distance field is the sum of the predicted results from $f_{base}$, $f_{fine}$, and $f_{norm}$.  Both $f_{base}$ and $f_{fine}$ are modeled by two MLPs, and the difference is that they adopt a stacked hourglass network and a shallow convolutional neural network as feature extractors respectively. This design aims to extract global features for large-scale shape recovery and to extract local features for middle-scale shape recovery. The two implicit functions are formulated as:

\begin{equation}
f_{base}: F(x, y), z \rightarrow d
\end{equation}
\begin{equation}
f_{fine}: F(x, y), z \rightarrow \nabla d = d - (d\otimes g_k)
\end{equation}

\noindent where $d$ is the signed distance sampled in the 3D volume; $F(x, y)$ is the pixel-aligned features extracted from the images; $\otimes$ is the 3D convolution operation with and the $k^3$ mean kernel. $k$ is set to $5$ in all our experiments.

We observed that the SDF and other implicit functions fail to model the geometric details when representing a large-scale database, so an additional normal carving operation $f_{norm}$ is presented to model small-scale geometry in an explicit manner. The pix2pixHD~\cite{wang2018pix2pixHD} is used to regress frontal and back normal maps by training on ground-truth normal maps. The normal carving operation $f_{norm}$ transforms the normal map to a 3D displacement field, which is formulated as:

\vspace{-0.05in}
\begin{equation}
f_{norm}: d \rightarrow d + \mathcal{N}(x,y) \otimes G_s
\end{equation}

\noindent where $\mathcal{N}(x,y)$ is the regressed frontal or back normal maps, and $G_s$ is the $7\times7$ Sobel operator. $f_{norm}$ apply the front and back normal maps to the front and back half of the signed distance field respectively. 

Empirically, the three signed distance fields predicted by $f_{base}$, $f_{fine}$, and $f_{norm}$ are designed to focus on large-scale, middle-scale, and small-scale 3D geometry respectively. Experiments show that the combination leads to both visually plausible and quantitatively accurate 3D face reconstruction.

%% file: 4_exp.tex
\begin{table*}[h]
\centering
\caption{Quantitative evaluation on MICC Florence dataset.
%\glw{New data(no forehead)!}
}
\vspace{-0.1in}
\begin{tabular}{lcccccccccc}
\toprule
                            \multicolumn{1}{c}{Pose Angle $\rightarrow$}  &      & $0^\circ$ &                         &      & $30^\circ$ &                         &       & $60^\circ$ &                         &           \\
\multicolumn{1}{c|}{Method \ $\downarrow$} & {CD}   & {\,MNE\,}    & \multicolumn{1}{c|}{\, CR\,} & {\, CD}   & {\, MNE}     & \multicolumn{1}{c|}{\, CR\,} & {\, CD}    & {\, MNE}     & \multicolumn{1}{c|}{\, CR\,} & Succ. \\ \midrule
Ext3dFace~\cite{tran2018extreme}            & 3.33    & 0.114     & 96.0    & 3.74    & 0.129     & 73.3    & 5.71    & 0.150     & 53.7    & 88.8     \\
PRNet~\cite{feng2018joint}           & \cc2.53    & 0.119     & 98.1    & \cb2.34    & 0.114     & 97.8    & \ca2.19    & \ce0.124     & \cd98.1    & \ca100.0     \\
D3DFR~\cite{deng2019accurate}            & 3.07    & 0.119     & 91.2    & 4.09    & 0.122     & 89.5    & 7.06    & 0.154     & 84.4    & \ca100.0     \\
RingNet~\cite{sanyal2019learning}            & \ca2.12    & \cc0.100     & \cc99.7    & 3.27    & \cc0.102     & \cc99.7    & 6.98    & 0.176     & \cc98.7    & \ca100.0     \\
DFDN~\cite{chen2019photo}            & 4.28    & 0.107     & \ce99.3    & 5.70    & 0.118     & \cd99.3    & 23.34    & 0.245     & 82.4    & 91.2     \\
DF2Net~\cite{zeng2019df2net}            & 3.60    & 0.130     & 79.5    & 6.09    & 0.190     & 64.1    & 7.81    & 0.210     & 46.1    & 58.8     \\
UDL~\cite{chen2020self}           & 2.70    & 0.110     & 94.4    & 2.96    & 0.111     & 93.7    & 5.42    & 0.169     & 84.0    & 80.0     \\
FaceScape~\cite{yang2020facescape}            & 3.90    & 0.124     & 97.2    & 3.66    & 0.129     & 95.0    & 5.76    & 0.187     & 86.9    & 99.2     \\
MGCNet~\cite{shang2020self}           & 3.00    & \ca0.086     & 95.7    & \ce2.83    & \cb0.090     & 95.4    & \cc2.82    & \cb0.096     & 94.5    & \ca100.0     \\
3DDFA\_v2~\cite{guo2020towards}            & \ce2.56    & \cb0.088     & 97.9    & \ca2.19    & \ca0.086     & 97.7    & \cb2.27    & \ca0.091     & \ce98.0    & \ca100.0     \\
SADRNet~\cite{ruan2021sadrnet}            & 6.10    & 0.134     & \ce99.3    & 5.66    & 0.129     & \ce98.6    & 6.73    & 0.141     & 96.9    & \ca100.0     \\
LAP~\cite{zhang2021learning}           & 3.74    & 0.121     & 97.6    & 6.02    & 0.152     & 96.8    & 10.78    & 0.214     & 88.4    & 99.2     \\
DECA~\cite{feng2021learning}            & \cd2.55    & \ce0.107     & \ca100.0    & \ce2.83    & \ce0.108     & \ca100.0    & \cd4.27    & \cc0.110     & \ca100.0    & \ca100.0     \\
Ours             & \cb2.25    & \cd0.102     & \ca100.0    & \cc2.53    & \cd0.104     & \ca100.0    & \ce4.66    & \cd0.114     & \cb99.9    & \ca100.0     \\
\bottomrule
\end{tabular}
\label{tab:eval_micc}
\end{table*}

\begin{table*}[t]
\centering
\caption{Ablation Study on FaceScape-wild dataset}
\vspace{-0.1in}
\resizebox{\textwidth}{!}
{ 
\begin{tabular}{lccccccccccccc}
\toprule
                            \multicolumn{1}{c}{Pose Angle $\rightarrow$}
                                       & 
                            \multicolumn{3}{c}{$0^\circ-5^\circ$}             & \multicolumn{3}{c}{$5^\circ-30^\circ$}            & \multicolumn{3}{c}{$30^\circ-60^\circ$}            & \multicolumn{3}{c}{$60^\circ-90^\circ$}             &           \\
\multicolumn{1}{c|}{Method \ $\downarrow$} & \,CD\,   & \,MNE\,   & \multicolumn{1}{c|}{\,CR\,} & \,CD\,   & \,MNE\,   & \multicolumn{1}{c|}{\,CR\,} & \,CD\,   & \,MNE\,   & \multicolumn{1}{c|}{\,CR\,} & \,CD\,    & \,MNE\,   & \multicolumn{1}{c|}{\,CR\,} & Succ. \\ \midrule
PRNet~\cite{feng2018joint}           & 2.61    & 0.119     & \ce83.0    & 3.11    & 0.114     & \ce82.7    & 4.26    & \ce0.119     & \ce78.2    & \ca3.88    & \ce0.140     & \ce75.3    & \ca100.0     \\
PRNet-pseudo data             & \cd1.97    & \ce0.096     & 78.5    & \cd2.01    & \ce0.102     & 78.3    & \cd2.61    & 0.122     & 75.5    & 6.93       & 0.212    & 51.2     & \ca100.0\\
PIFu~\cite{saito2019pifu}             & \ce2.14    & \cd0.066     & \cb99.4    & \ce2.25    & \cd0.070     & \cb99.2    & \ce3.10    & \cd0.084     & \ca98.6    & \cb4.92       & \ca0.118    & \ca97.1     & \ca100.0\\
$f_{base}$             & \cc1.81    & \cb0.062     & \cc99.0    & \cc1.95    & \cb0.066     & \cc98.8    & \cc2.16    & \cb0.072     & \cb98.4    & \cc5.04       & \cc0.124    & \cc95.7     & \ca100.0\\
$f_{base}+f_{fine}$  & \ca1.79    & \cb0.062     & \cd98.9    & \ca1.90    & \cb0.066     & \cc98.8    & \ca2.12    & \cc0.073     & \cb98.4    & \cd5.16    & \cd0.125    & \cc95.7     & \ca100.0\\
$f_{base}+f_{fine}+f_{norm}$  & \ca1.79    & \ca0.058     & \ca99.5    & \cb1.92    & \ca0.062     & \ca99.3    & \ca2.12    & \ca0.071     & \cb98.4    & \ce5.26    & \cb0.123    & \cb96.7     & \ca100.0\\

\bottomrule
\end{tabular}
}
\label{tab:ablation}
\end{table*}

\section{Experiment}
\label{sec:exp}

\subsection{Dataset and Metric}
The qualitative evaluation is on FFHQ~\cite{ffhq}, IMDB-WIKI~\cite{rothe2015dex}, and AFLW2000~\cite{zhu2016face} datasets, which are in-the-wild 2D image datasets covering different poses, races, ages, environments, and lighting. The quantitative evaluation is on FaceScape~\cite{yang2020facescape,zhu2021facescape}, and MICC Florence~\cite{micc}, which are face datasets containing accurate 3D shapes and 2D images. 

We follow the evaluation methodology explained in FaceScape benchmark\cite{yang2020facescape, zhu2021facescape} to evaluate the accuracy of the reconstructed shape at the time of the photo, which means poses and expressions are also factored into the calculation of error. Generally, CD measures the overall shape accuracy; MNE measures the local shape accuracy; CR indicates if the result face is complete. 
Different from NoW benchmarks \cite{RingNet:CVPR:2019}, which evaluate the `expression-neutralized' and `unposed' facial shape, the benchmark we used evaluates the accuracy of the predicted 3D shape at the `time of the photo' and takes pose estimation into consideration.

\subsection{Visual Comparison}
\begin{figure*}[tb]
\begin{center}
\includegraphics[width=1.0\linewidth]{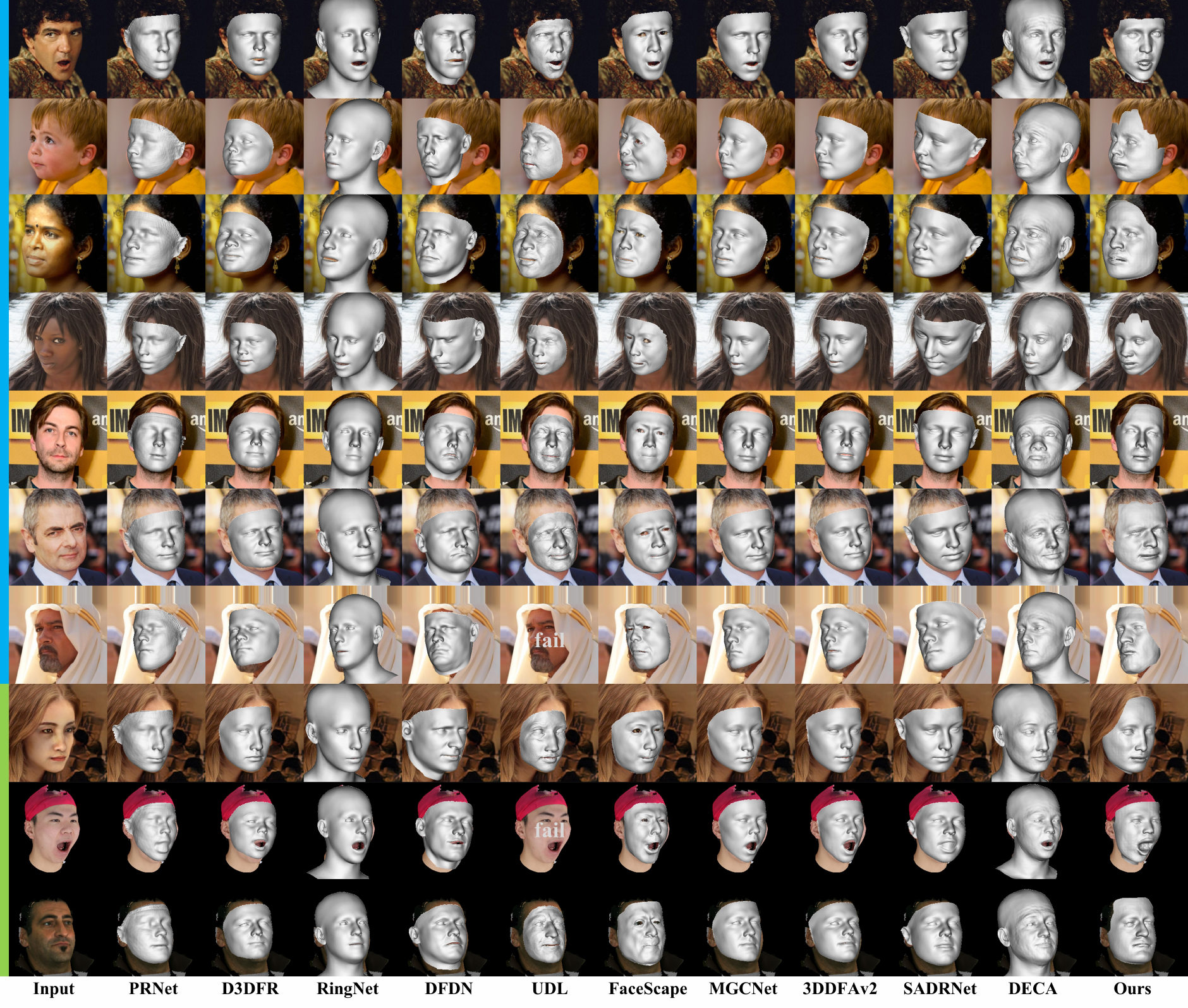}
\end{center}
\vspace{-0.2in}
\caption{
Qualitative comparison. Images with blue bars are from in-the-wild datasets(FFHQ, IMDB-WIKI, AFLW2000); images with green bars are from synthetic or studio-captured datasets (FaceScape-wild/lab, MICC-Florence). 
}
\label{fig:visual}
\vspace{-0.1in}
\end{figure*}

We show our results on extreme conditions in Figure~\ref{fig:show} and compare our method with previous methods in Figure~\ref{fig:visual}.  We can see that our results are visually more faithful and well-aligned with the source image. We believe it is because our created datasets eliminate the gaps between 2D images and 3D shapes and boost the performance of the non-parametric SVFR model. Please watch our video to validate the temporal continuity.

\subsection{Quantitative Comparison}

We conduct quantitative evaluations on three datasets. The evaluation on FaceScape-wild \cite{zhu2021facescape} dataset is reported in Table~\ref{tab:eval_wild}. FaceScape-wild dataset contains $400$ synthetic in-the-wild images categorized by the pose angle. We can see that our method leads in CD and MNE when pose angles are in $0^{\circ}-60^{\circ}$ angles. Our performance in $60^{\circ}-90^{\circ}$ angles are slightly worse but still ranks top 5 in all metrics. Our method is the only one that ranks the top 5 in all metrics and leads in the $0^{\circ}-60^{\circ}$ pose by a large margin. 

The quantitative evaluation on FaceScape-lab \cite{zhu2021facescape} dataset is reported in Table~\ref{tab:eval_lab}. FaceScape-lab dataset contains $660$ images rendered from $20$ studio-captured detailed 3D facial raw scans using the perspective projection camera model. 
Similarly, our method is the only one that is in the top five across all metrics. 

The quantitative evaluation on MICC Florence~\cite{micc} dataset is reported in Table~\ref{tab:eval_lab}.  MICC Florence dataset contains $52$ studio-captured detailed 3D face models, which are rendered to $260$ images using the perspective projection camera model. 
The major difference compared with FaceScape-lab is that MICC Florence mainly contains western faces while FaceScape-lab contains oriental faces. 
Our method leads on MICC Florence benchmark in the mean of average ranking of all metrics, which shows that our models work robustly and accurately for both western and oriental faces.

\subsection{Ablation Study}

The results of the ablation study are shown in Table~\ref{tab:ablation}. `PRNet' is the model that is trained on the original fitted dataset; `PRNet-pseudo data' means the model trained on our pseudo 3D\&3D data; `PIFu' is a non-parametric single-view reconstruction method trained on our 2D\&3D data; `$f_{base}$', `$f_{base}+f_{fine}$' and `$f_{base}+f_{fine}+f_{norm}$' means part of our hierarchical SDF as explained in Figure~\ref{fig:hsdf}. 

By comparing `PRNet' and `PRNet-pseudo data', we can conclude that our pseudo 2D\&3D data can be used in improving other non-parametric methods. By comparing `PIFu' to `$f_{base}$', we verify that introducing SDF improves the performance of SVFR. By comparing `$f_{base}$', `$f_{base}+f_{fine}$' and `$f_{base}+f_{fine}+f_{norm}$', we verify that the hierarchical SDF further improves performance in each level. $f_{fine}$ focuses on improving the middle-scale shape and improving the accuracy in terms of CD. Though the additional $f_{norm}$ doesn't affect much to CD, it enhances the MNE which means the detailed shape is recovered more accurately. These improvements are also confirmed by visual comparison in Figure~\ref{fig:hsdf}. For the viewing angle $>60^{\circ}$ where almost half of the facial regions are occluded, $f_{fine}$ and $f_{norm}$ struggled in hallucinating uncertain occluded shape, thus the score of CD reduces slightly.

\subsection{Discussion of Limitations.}

\textbf{Non-uniform mesh topology.} Due to the non-parametric scheme, the topology of our result meshes is not uniform, so an additional registration phase is required before being used in downstream applications like rigging and animation. The non-uniform topology also increases the possibility to produce broken mesh in extreme poses and lighting conditions, which will be shown in the supplementary.

\noindent\textbf{Inveracious lightings.} In some of our pseudo images, the synthesized lighting is not photo-realistic enough with no self-shading and specular highlight. Besides, the illuminations on the face and the background may differ much in some cases. Whether these factors will affect performance remains an open question.

\noindent\textbf{Inferior performance under large poses.} From the quantitative evaluation (Table~\ref{tab:eval_wild}/\ref{tab:eval_lab}/\ref{tab:eval_micc}), we can see that the reconstruction accuracy of minor-pose faces in our method is significantly higher than that of large-pose faces. We believe that the key reason is the insufficient large-pose wild images used in generating our pseudo 2D\&3D data. Specifically, only $0.18\%$ of the photos in CelebA dataset and $0.21\%$ of the photos in the FFHQ dataset are taken from views at $60^\circ-90^\circ$. We manually add more large-pose images from other datasets, increasing the ratio of large-pose images in our training set to $12.14\%$, but they are still too few. This problem is not unique to our method but to most previous methods, as illustrated in Table~\ref{tab:eval_wild}/\ref{tab:eval_lab}/\ref{tab:eval_micc}. We consider that how to augment the data with large-pose faces remains a challenge for the SVFR task.

%% file: 5_con.tex
\vspace{0.08in}
\section{Conclusion}
\label{sec:con}

We propose a novel approach for the single-view 3D face reconstruction task in a non-parametric scheme. Our method gets rid of the heavy dependence on the statistic model and, therefore, its limitations and achieves state-of-the-art performance by learning from our created pseudo 2D\&3D datasets. A novel solution to build a large-scale and accurate in-the-wild 3D face dataset is presented, filling the gap of image-shape alignment in previous datasets. We hope our work could inspire future researchers in this field.

%% file: 0_main.bbl
\begin{thebibliography}{61}
\providecommand{\natexlab}[1]{#1}

\bibitem[{Amberg, Romdhani, and Vetter(2007)}]{amberg2007optimal}
Amberg, B.; Romdhani, S.; and Vetter, T. 2007.
\newblock Optimal step nonrigid ICP algorithms for surface registration.
\newblock In \emph{CVPR}, 1--8.

\bibitem[{Bagdanov, Del~Bimbo, and Masi(2011)}]{micc}
Bagdanov, A.~D.; Del~Bimbo, A.; and Masi, I. 2011.
\newblock The Florence 2D/3D Hybrid Face Dataset.
\newblock In \emph{Joint ACM Workshop on Human Gesture and Behavior
  Understanding}, 79–80.

\bibitem[{Blanz, Vetter et~al.(1999)}]{blanz1999morphable}
Blanz, V.; Vetter, T.; et~al. 1999.
\newblock A morphable model for the synthesis of 3D faces.
\newblock In \emph{SIGGRAPH}, volume~99, 187--194.

\bibitem[{Cao et~al.(2013{\natexlab{a}})Cao, Weng, Lin, and Zhou}]{cao20133d}
Cao, C.; Weng, Y.; Lin, S.; and Zhou, K. 2013{\natexlab{a}}.
\newblock 3D shape regression for real-time facial animation.
\newblock \emph{TOG}, 32(4): 1--10.

\bibitem[{Cao et~al.(2013{\natexlab{b}})Cao, Weng, Zhou, Tong, and
  Zhou}]{cao2013facewarehouse}
Cao, C.; Weng, Y.; Zhou, S.; Tong, Y.; and Zhou, K. 2013{\natexlab{b}}.
\newblock Facewarehouse: A 3d facial expression database for visual computing.
\newblock \emph{TVCG}, 20(3): 413--425.

\bibitem[{Chen et~al.(2019)Chen, Chen, Zhang, Mitchell, and Yu}]{chen2019photo}
Chen, A.; Chen, Z.; Zhang, G.; Mitchell, K.; and Yu, J. 2019.
\newblock Photo-realistic facial details synthesis from single image.
\newblock In \emph{ICCV}, 9429--9439.

\bibitem[{Chen et~al.(2020)Chen, Wu, Wang, Song, Ling, and Bao}]{chen2020self}
Chen, Y.; Wu, F.; Wang, Z.; Song, Y.; Ling, Y.; and Bao, L. 2020.
\newblock Self-supervised learning of detailed 3d face reconstruction.
\newblock \emph{TIP}, 29: 8696--8705.

\bibitem[{Dai et~al.(2019)Dai, Pears, Smith, and Duncan}]{Dai2019}
Dai, H.; Pears, N.; Smith, W.; and Duncan, C. 2019.
\newblock Statistical Modeling of Craniofacial Shape and Texture.
\newblock \emph{IJCV}, 128(2): 547--571.

\bibitem[{Deng et~al.(2019)Deng, Yang, Xu, Chen, Jia, and
  Tong}]{deng2019accurate}
Deng, Y.; Yang, J.; Xu, S.; Chen, D.; Jia, Y.; and Tong, X. 2019.
\newblock Accurate 3d face reconstruction with weakly-supervised learning: From
  single image to image set.
\newblock In \emph{CVPRW}, 0--0.

\bibitem[{Dou, Shah, and Kakadiaris(2017)}]{dou2017end}
Dou, P.; Shah, S.~K.; and Kakadiaris, I.~A. 2017.
\newblock End-to-end 3D face reconstruction with deep neural networks.
\newblock In \emph{CVPR}, 5908--5917.

\bibitem[{Egger et~al.(2020)Egger, Smith, Tewari, Wuhrer, Zollhoefer, Beeler,
  Bernard, Bolkart, Kortylewski, Romdhani et~al.}]{egger20203d}
Egger, B.; Smith, W.~A.; Tewari, A.; Wuhrer, S.; Zollhoefer, M.; Beeler, T.;
  Bernard, F.; Bolkart, T.; Kortylewski, A.; Romdhani, S.; et~al. 2020.
\newblock 3d morphable face models—past, present, and future.
\newblock \emph{TOG}, 39(5): 1--38.

\bibitem[{Feng et~al.(2021)Feng, Feng, Black, and Bolkart}]{feng2021learning}
Feng, Y.; Feng, H.; Black, M.~J.; and Bolkart, T. 2021.
\newblock Learning an animatable detailed 3D face model from in-the-wild
  images.
\newblock \emph{TOG}, 40(4): 1--13.

\bibitem[{Feng et~al.(2018)Feng, Wu, Shao, Wang, and Zhou}]{feng2018joint}
Feng, Y.; Wu, F.; Shao, X.; Wang, Y.; and Zhou, X. 2018.
\newblock Joint 3d face reconstruction and dense alignment with position map
  regression network.
\newblock In \emph{ECCV}, 534--551.

\bibitem[{Gecer et~al.(2019)Gecer, Ploumpis, Kotsia, and
  Zafeiriou}]{gecer2019ganfit}
Gecer, B.; Ploumpis, S.; Kotsia, I.; and Zafeiriou, S. 2019.
\newblock Ganfit: Generative adversarial network fitting for high fidelity 3d
  face reconstruction.
\newblock In \emph{CVPR}, 1155--1164.

\bibitem[{Genova et~al.(2018)Genova, Cole, Maschinot, Sarna, Vlasic, and
  Freeman}]{genova2018unsupervised}
Genova, K.; Cole, F.; Maschinot, A.; Sarna, A.; Vlasic, D.; and Freeman, W.~T.
  2018.
\newblock Unsupervised training for 3d morphable model regression.
\newblock In \emph{CVPR}, 8377--8386.

\bibitem[{Guo et~al.(2020)Guo, Zhu, Yang, Yang, Lei, and Li}]{guo2020towards}
Guo, J.; Zhu, X.; Yang, Y.; Yang, F.; Lei, Z.; and Li, S.~Z. 2020.
\newblock Towards fast, accurate and stable 3d dense face alignment.
\newblock In \emph{ECCV}, 152--168.

\bibitem[{Jackson et~al.(2017)Jackson, Bulat, Argyriou, and
  Tzimiropoulos}]{jackson2017large}
Jackson, A.~S.; Bulat, A.; Argyriou, V.; and Tzimiropoulos, G. 2017.
\newblock Large pose 3D face reconstruction from a single image via direct
  volumetric CNN regression.
\newblock In \emph{ICCV}, 1031--1039.

\bibitem[{Jiang et~al.(2018)Jiang, Zhang, Deng, Li, and Liu}]{jiang20183d}
Jiang, L.; Zhang, J.; Deng, B.; Li, H.; and Liu, L. 2018.
\newblock 3D face reconstruction with geometry details from a single image.
\newblock \emph{TIP}, 27(10): 4756--4770.

\bibitem[{Karras, Laine, and Aila(2019)}]{ffhq}
Karras, T.; Laine, S.; and Aila, T. 2019.
\newblock A style-based generator architecture for generative adversarial
  networks.
\newblock In \emph{CVPR}, 4401--4410.

\bibitem[{Koizumi and Smith(2020)}]{koizumi2020look}
Koizumi, T.; and Smith, W.~A. 2020.
\newblock “Look Ma, no landmarks!”--Unsupervised, model-based dense face
  alignment.
\newblock In \emph{ECCV}, 690--706. Springer.

\bibitem[{Lee et~al.(2020)Lee, Liu, Wu, and Luo}]{lee2020maskgan}
Lee, C.-H.; Liu, Z.; Wu, L.; and Luo, P. 2020.
\newblock Maskgan: Towards diverse and interactive facial image manipulation.
\newblock In \emph{CVPR}, 5549--5558.

\bibitem[{Li et~al.(2017)Li, Bolkart, Black, Li, and Romero}]{li2017learning}
Li, T.; Bolkart, T.; Black, M.~J.; Li, H.; and Romero, J. 2017.
\newblock Learning a model of facial shape and expression from 4D scans.
\newblock \emph{TOG}, 36(6): 194--1.

\bibitem[{Liu et~al.(2018)Liu, Zhu, Zeng, Zhao, and Liu}]{liu2018disentangling}
Liu, F.; Zhu, R.; Zeng, D.; Zhao, Q.; and Liu, X. 2018.
\newblock Disentangling features in 3D face shapes for joint face
  reconstruction and recognition.
\newblock In \emph{CVPR}, 5216--5225.

\bibitem[{Richardson, Sela, and Kimmel(2016)}]{richardson20163d}
Richardson, E.; Sela, M.; and Kimmel, R. 2016.
\newblock 3D face reconstruction by learning from synthetic data.
\newblock In \emph{3DV}, 460--469. IEEE.

\bibitem[{Richardson et~al.(2017)Richardson, Sela, Orel, and
  Kimmel}]{richardson2017learning}
Richardson, E.; Sela, M.; Orel, R.; and Kimmel, R. 2017.
\newblock Learning Detailed Face Reconstruction from a Single Image.
\newblock In \emph{CVPR}, 5553--5562.

\bibitem[{Riviere et~al.(2020)Riviere, Gotardo, Bradley, Ghosh, and
  Beeler}]{riviere2020single}
Riviere, J.; Gotardo, P.; Bradley, D.; Ghosh, A.; and Beeler, T. 2020.
\newblock Single-shot high-quality facial geometry and skin appearance capture.
\newblock \emph{TOG}, 39(4): 81--1.

\bibitem[{Romdhani and Vetter(2005)}]{romdhani2005estimating}
Romdhani, S.; and Vetter, T. 2005.
\newblock Estimating 3D shape and texture using pixel intensity, edges,
  specular highlights, texture constraints and a prior.
\newblock In \emph{CVPR}, volume~2, 986--993.

\bibitem[{Rothe, Timofte, and Van~Gool(2015)}]{rothe2015dex}
Rothe, R.; Timofte, R.; and Van~Gool, L. 2015.
\newblock Dex: Deep expectation of apparent age from a single image.
\newblock In \emph{ICCV Workshops}, 10--15.

\bibitem[{Ruan et~al.(2021)Ruan, Zou, Wu, Wu, and Wang}]{ruan2021sadrnet}
Ruan, Z.; Zou, C.; Wu, L.; Wu, G.; and Wang, L. 2021.
\newblock SADRNet: Self-Aligned Dual Face Regression Networks for Robust 3D
  Dense Face Alignment and Reconstruction.
\newblock \emph{TIP}.

\bibitem[{Sagonas et~al.(2016)Sagonas, Antonakos, Tzimiropoulos, Zafeiriou, and
  Pantic}]{sagonas2016300}
Sagonas, C.; Antonakos, E.; Tzimiropoulos, G.; Zafeiriou, S.; and Pantic, M.
  2016.
\newblock 300 faces in-the-wild challenge: Database and results.
\newblock \emph{Image and Vision Computing}, 47: 3--18.

\bibitem[{Saito et~al.(2019)Saito, Huang, Natsume, Morishima, Kanazawa, and
  Li}]{saito2019pifu}
Saito, S.; Huang, Z.; Natsume, R.; Morishima, S.; Kanazawa, A.; and Li, H.
  2019.
\newblock Pifu: Pixel-aligned implicit function for high-resolution clothed
  human digitization.
\newblock In \emph{ICCV}, 2304--2314.

\bibitem[{Sanyal et~al.(2019{\natexlab{a}})Sanyal, Bolkart, Feng, and
  Black}]{RingNet:CVPR:2019}
Sanyal, S.; Bolkart, T.; Feng, H.; and Black, M. 2019{\natexlab{a}}.
\newblock Learning to Regress 3D Face Shape and Expression from an Image
  without 3D Supervision.
\newblock In \emph{CVPR}.

\bibitem[{Sanyal et~al.(2019{\natexlab{b}})Sanyal, Bolkart, Feng, and
  Black}]{sanyal2019learning}
Sanyal, S.; Bolkart, T.; Feng, H.; and Black, M.~J. 2019{\natexlab{b}}.
\newblock Learning to regress 3D face shape and expression from an image
  without 3D supervision.
\newblock In \emph{CVPR}, 7763--7772.

\bibitem[{Sela, Richardson, and Kimmel(2017)}]{sela2017unrestricted}
Sela, M.; Richardson, E.; and Kimmel, R. 2017.
\newblock Unrestricted facial geometry reconstruction using image-to-image
  translation.
\newblock In \emph{ICCV}, 1576--1585.

\bibitem[{Sengupta et~al.(2018)Sengupta, Kanazawa, Castillo, and
  Jacobs}]{sengupta2018sfsnet}
Sengupta, S.; Kanazawa, A.; Castillo, C.~D.; and Jacobs, D.~W. 2018.
\newblock SfSNet: Learning Shape, Reflectance and Illuminance of Faces in the
  Wild.
\newblock In \emph{CVPR}, 6296--6305.

\bibitem[{Shang et~al.(2020)Shang, Shen, Li, Zhou, Zhen, Fang, and
  Quan}]{shang2020self}
Shang, J.; Shen, T.; Li, S.; Zhou, L.; Zhen, M.; Fang, T.; and Quan, L. 2020.
\newblock Self-supervised monocular 3d face reconstruction by occlusion-aware
  multi-view geometry consistency.
\newblock In \emph{ECCV}, 53--70.

\bibitem[{Tan et~al.(2020)Tan, Zhu, Cui, Zhu, Pollefeys, and Tan}]{tan2020self}
Tan, F.; Zhu, H.; Cui, Z.; Zhu, S.; Pollefeys, M.; and Tan, P. 2020.
\newblock Self-supervised human depth estimation from monocular videos.
\newblock In \emph{CVPR}, 650--659.

\bibitem[{Tewari et~al.(2018)Tewari, Zollh{\"o}fer, Garrido, Bernard, Kim,
  P{\'e}rez, and Theobalt}]{tewari2018self}
Tewari, A.; Zollh{\"o}fer, M.; Garrido, P.; Bernard, F.; Kim, H.; P{\'e}rez,
  P.; and Theobalt, C. 2018.
\newblock Self-supervised multi-level face model learning for monocular
  reconstruction at over 250 hz.
\newblock In \emph{CVPR}, 2549--2559.

\bibitem[{Tewari et~al.(2017)Tewari, Zollhofer, Kim, Garrido, Bernard, Perez,
  and Theobalt}]{tewari2017mofa}
Tewari, A.; Zollhofer, M.; Kim, H.; Garrido, P.; Bernard, F.; Perez, P.; and
  Theobalt, C. 2017.
\newblock Mofa: Model-based deep convolutional face autoencoder for
  unsupervised monocular reconstruction.
\newblock In \emph{CVPRW}, 1274--1283.

\bibitem[{Thies et~al.(2016)Thies, Zollhofer, Stamminger, Theobalt, and
  Nie{\ss}ner}]{thies2016face2face}
Thies, J.; Zollhofer, M.; Stamminger, M.; Theobalt, C.; and Nie{\ss}ner, M.
  2016.
\newblock Face2face: Real-time face capture and reenactment of rgb videos.
\newblock In \emph{CVPR}, 2387--2395.

\bibitem[{Tran et~al.(2018)Tran, Hassner, Masi, Paz, Nirkin, and
  Medioni}]{tran2018extreme}
Tran, A.~T.; Hassner, T.; Masi, I.; Paz, E.; Nirkin, Y.; and Medioni, G.~G.
  2018.
\newblock Extreme 3D Face Reconstruction: Seeing Through Occlusions.
\newblock In \emph{CVPR}, 3935--3944.

\bibitem[{Tran, Liu, and Liu(2019)}]{tran2019towards}
Tran, L.; Liu, F.; and Liu, X. 2019.
\newblock Towards high-fidelity nonlinear 3D face morphable model.
\newblock In \emph{CVPR}, 1126--1135.

\bibitem[{Tran and Liu(2018)}]{tran2018nonlinear}
Tran, L.; and Liu, X. 2018.
\newblock Nonlinear 3D face morphable model.
\newblock In \emph{CVPR}, 7346--7355.

\bibitem[{Tu et~al.(2020)Tu, Zhao, Xie, Jiang, Balamurugan, Luo, Zhao, He, Ma,
  and Feng}]{tu20203d}
Tu, X.; Zhao, J.; Xie, M.; Jiang, Z.; Balamurugan, A.; Luo, Y.; Zhao, Y.; He,
  L.; Ma, Z.; and Feng, J. 2020.
\newblock 3d face reconstruction from a single image assisted by 2d face images
  in the wild.
\newblock \emph{TMM}, 23: 1160--1172.

\bibitem[{Varol et~al.(2017)Varol, Romero, Martin, Mahmood, Black, Laptev, and
  Schmid}]{varol2017learning}
Varol, G.; Romero, J.; Martin, X.; Mahmood, N.; Black, M.~J.; Laptev, I.; and
  Schmid, C. 2017.
\newblock Learning from synthetic humans.
\newblock In \emph{CVPR}, 109--117.

\bibitem[{Wang et~al.(2018)Wang, Liu, Zhu, Tao, Kautz, and
  Catanzaro}]{wang2018pix2pixHD}
Wang, T.-C.; Liu, M.-Y.; Zhu, J.-Y.; Tao, A.; Kautz, J.; and Catanzaro, B.
  2018.
\newblock High-Resolution Image Synthesis and Semantic Manipulation with
  Conditional GANs.
\newblock In \emph{CVPR}.

\bibitem[{Xiao et~al.(2022)Xiao, Zhu, Yang, Diao, Lu, and
  Cao}]{xiao2022detailed}
Xiao, Y.; Zhu, H.; Yang, H.; Diao, Z.; Lu, X.; and Cao, X. 2022.
\newblock Detailed Facial Geometry Recovery from Multi-view Images by Learning
  an Implicit Function.
\newblock In \emph{AAAI}.

\bibitem[{Yang et~al.(2020)Yang, Zhu, Wang, Huang, Shen, Yang, and
  Cao}]{yang2020facescape}
Yang, H.; Zhu, H.; Wang, Y.; Huang, M.; Shen, Q.; Yang, R.; and Cao, X. 2020.
\newblock FaceScape: A Large-Scale High Quality 3D Face Dataset and Detailed
  Riggable 3D Face Prediction.
\newblock In \emph{CVPR}.

\bibitem[{Yenamandra et~al.(2021)Yenamandra, Tewari, Bernard, Seidel, Elgharib,
  Cremers, and Theobalt}]{yenamandra2021i3dmm}
Yenamandra, T.; Tewari, A.; Bernard, F.; Seidel, H.-P.; Elgharib, M.; Cremers,
  D.; and Theobalt, C. 2021.
\newblock i3dmm: Deep implicit 3d morphable model of human heads.
\newblock In \emph{CVPR}, 12803--12813.

\bibitem[{Yu et~al.(2018)Yu, Wang, Peng, Gao, Yu, and Sang}]{yu2018bisenet}
Yu, C.; Wang, J.; Peng, C.; Gao, C.; Yu, G.; and Sang, N. 2018.
\newblock Bisenet: Bilateral segmentation network for real-time semantic
  segmentation.
\newblock In \emph{ECCV}.

\bibitem[{Zeng, Peng, and Qiao(2019)}]{zeng2019df2net}
Zeng, X.; Peng, X.; and Qiao, Y. 2019.
\newblock DF2Net: A dense-fine-finer network for detailed 3D face
  reconstruction.
\newblock In \emph{ICCV}, 2315--2324.

\bibitem[{Zhang et~al.(2021)Zhang, Ge, Chen, Tai, Yan, Yang, Wang, Li, and
  Huang}]{zhang2021learning}
Zhang, Z.; Ge, Y.; Chen, R.; Tai, Y.; Yan, Y.; Yang, J.; Wang, C.; Li, J.; and
  Huang, F. 2021.
\newblock Learning to Aggregate and Personalize 3D Face from In-the-Wild Photo
  Collection.
\newblock In \emph{CVPR}, 14214--14224.

\bibitem[{Zhou et~al.(2019)Zhou, Deng, Kotsia, and Zafeiriou}]{zhou2019dense}
Zhou, Y.; Deng, J.; Kotsia, I.; and Zafeiriou, S. 2019.
\newblock Dense 3d face decoding over 2500fps: Joint texture \& shape
  convolutional mesh decoders.
\newblock In \emph{CVPR}, 1097--1106.

\bibitem[{Zhu et~al.(2018)Zhu, Su, Wang, Cao, and Yang}]{zhu2018view}
Zhu, H.; Su, H.; Wang, P.; Cao, X.; and Yang, R. 2018.
\newblock View extrapolation of human body from a single image.
\newblock In \emph{CVPR}, 4450--4459.

\bibitem[{Zhu et~al.(2021{\natexlab{a}})Zhu, Yang, Guo, Zhang, Wang, Huang,
  Shen, Yang, and Cao}]{zhu2021facescape}
Zhu, H.; Yang, H.; Guo, L.; Zhang, Y.; Wang, Y.; Huang, M.; Shen, Q.; Yang, R.;
  and Cao, X. 2021{\natexlab{a}}.
\newblock FaceScape: 3D Facial Dataset and Benchmark for Single-View 3D Face
  Reconstruction.
\newblock \emph{arXiv preprint arXiv:2111.01082}.

\bibitem[{Zhu et~al.(2019)Zhu, Zuo, Wang, Cao, and Yang}]{zhu2019detailed}
Zhu, H.; Zuo, X.; Wang, S.; Cao, X.; and Yang, R. 2019.
\newblock Detailed human shape estimation from a single image by hierarchical
  mesh deformation.
\newblock In \emph{CVPR}, 4491--4500.

\bibitem[{Zhu et~al.(2021{\natexlab{b}})Zhu, Zuo, Yang, Wang, Cao, and
  Yang}]{zhu2021detailed}
Zhu, H.; Zuo, X.; Yang, H.; Wang, S.; Cao, X.; and Yang, R. 2021{\natexlab{b}}.
\newblock Detailed avatar recovery from single image.
\newblock \emph{PAMI}.

\bibitem[{Zhu et~al.(2016)Zhu, Lei, Liu, Shi, and Li}]{zhu2016face}
Zhu, X.; Lei, Z.; Liu, X.; Shi, H.; and Li, S.~Z. 2016.
\newblock Face alignment across large poses: A 3d solution.
\newblock In \emph{CVPR}, 146--155.

\bibitem[{Zhu et~al.(2020)Zhu, Yang, Huang, Yu, Wang, Guo, Lei, and
  Li}]{zhu2020beyond}
Zhu, X.; Yang, F.; Huang, D.; Yu, C.; Wang, H.; Guo, J.; Lei, Z.; and Li, S.~Z.
  2020.
\newblock Beyond 3dmm space: Towards fine-grained 3d face reconstruction.
\newblock In \emph{ECCV}, 343--358.

\bibitem[{Zhuang et~al.(2022)Zhuang, Zhu, Sun, and Cao}]{zhuang2022mofanerf}
Zhuang, Y.; Zhu, H.; Sun, X.; and Cao, X. 2022.
\newblock Mofanerf: Morphable facial neural radiance field.
\newblock In \emph{ECCV}, 268--285. Springer.

\bibitem[{Zollh{\"o}fer et~al.(2018)Zollh{\"o}fer, Thies, Garrido, Bradley,
  Beeler, P{\'e}rez, Stamminger, Nie{\ss}ner, and
  Theobalt}]{zollhofer2018state}
Zollh{\"o}fer, M.; Thies, J.; Garrido, P.; Bradley, D.; Beeler, T.; P{\'e}rez,
  P.; Stamminger, M.; Nie{\ss}ner, M.; and Theobalt, C. 2018.
\newblock State of the Art on Monocular 3D Face Reconstruction, Tracking, and
  Applications.
\newblock In \emph{CGF}, volume~37, 523--550.

\end{thebibliography}
